\documentclass[10pt,twocolumn,letterpaper]{article}

\usepackage{cvpr}   
\usepackage{multicol}
\usepackage{amssymb}
\usepackage{pifont}
\usepackage{tabularx}

\newcommand{\cmark}{\ding{51}} 
\newcommand{\xmark}{\ding{55}} 

\usepackage{graphicx}
\usepackage{amsmath}
\usepackage{amssymb}
\usepackage{booktabs}
\usepackage{multirow}
\usepackage{makecell}
\usepackage{soul}
\usepackage[table,dvipsnames]{xcolor}
\usepackage{changepage,threeparttable}
\usepackage[accsupp]{axessibility}
\usepackage{mmstyle}

\definecolor{MyRed}{HTML}{FF0000}

\usepackage[misc]{ifsym}

\usepackage{pifont}

\definecolor{cvprblue}{rgb}{0.21,0.49,0.74}
\usepackage[pagebackref,breaklinks,colorlinks,citecolor=cvprblue]{hyperref}

\usepackage[capitalize]{cleveref}
\crefname{section}{Sec.}{Secs.}
\Crefname{section}{Section}{Sections}
\Crefname{table}{Table}{Tables}
\crefname{table}{Tab.}{Tabs.}

\makeatletter
\DeclareRobustCommand\onedot{\futurelet\@let@token\@onedot}
\def\@onedot{\ifx\@let@token.\else.\null\fi\xspace}

\makeatother

\definecolor{turquoise}{cmyk}{0.65,0,0.1,0.3}
\definecolor{purple}{rgb}{0.65,0,0.65}
\definecolor{dark_green}{rgb}{0, 0.5, 0}
\definecolor{orange}{rgb}{0.8, 0.6, 0.2}
\definecolor{red}{rgb}{0.8, 0.2, 0.2}
\definecolor{darkred}{rgb}{0.6, 0.1, 0.05}
\definecolor{blueish}{rgb}{0.0, 0.3, .6}
\definecolor{light_gray}{rgb}{0.7, 0.7, .7}
\definecolor{pink}{rgb}{1, 0, 1}
\definecolor{greyblue}{rgb}{0.25, 0.25, 1}
\definecolor{colorFst}{HTML}{bde6cd}       %
\definecolor{colorSnd}{HTML}{e4eebc}       %
\definecolor{colorTrd}{HTML}{fff8c5}       %

\def\paperID{4} %
\def\confName{CVPR}
\def\confYear{2024}

\title{MARINE: A Computer Vision Model for Detecting Rare Predator-Prey Interactions in Animal Videos}

\author{Zsófia Katona \and
Seyed Sahand Mohammadi Ziabari \and
Fatemeh Karimi Nejadasl \and
University of Amsterdam}

\begin{document}
\maketitle
\begin{abstract}
Encounters between predator and prey play an essential role in ecosystems, but their rarity makes them difficult to detect in video recordings. Although advances in action recognition (AR) and temporal action detection (AD), especially transformer-based models and vision foundation models, have achieved high performance on human action datasets, animal videos remain relatively under-researched. This thesis addresses this gap by proposing the model MARINE, which utilizes motion-based frame selection designed for fast animal actions and DINOv2 feature extraction with a trainable classification head for action recognition. MARINE outperforms VideoMAE in identifying predator attacks in videos of fish, both on a small and specific coral reef dataset (81.53\% against 52.64\% accuracy), and on a subset of the more extensive Animal Kingdom dataset (94.86\% against 83.14\% accuracy). In a multi-label setting on a representative sample of Animal Kingdom, MARINE achieves 23.79\% mAP, positioning it mid-field among existing benchmarks. Furthermore, in an AD task on the coral reef dataset, MARINE achieves 80.78\% AP (against VideoMAE’s 34.89\%) although at a lowered t-IoU threshold of 25\%. Therefore, despite room for improvement, MARINE offers an effective starter framework to apply to AR and AD tasks on animal recordings and thus contribute to the study of natural ecosystems.
\footnote{https://github.com/ZsofiaK/masterthesis}
\footnote{This is an MSc thesis by Zsófia Katona, supervised by the two other authors.}
\footnote{The original title was: The automatic detection of rare events in videos of wild animals: Introducing MARINE, a computer vision model trained on fish predator attacks}
\end{abstract}
    
\section{Introduction}
\label{sec:introduction}




Encounters between predator and prey play an essential role in ecosystem dynamics. While it has been long understood that predators  play a crucial role in the regulation of prey populations, the last decades have shown a growing interest in the complexities of predator-prey interactions and their importance in terms of the ecosystem and biodiversity \cite{ives_synthesis_2005}. An understanding of predator-prey interactions is especially important in light of the ongoing climate change, as a changing climate has been shown to alter predator-prey dynamics across a range of different habitats and species \cite{laws_climate_2017,sadykova_ecological_2020, damien_preypredator_2019}, and it is also implied that the understanding of predator-prey interactions is crucial in designing sustainable solutions in biological systems \cite{schmitz_climate_2014,hunsicker_functional_2011}.

At the same time, the observation of predator-prey interactions presents a significant challenge even in a time when video footage from wildlife camera traps is available. The reason behind this is the rarity of these interaction events (such as attacks), which means that relevant sections of the camera footage typically comprise only an extremely small fraction of the total length of videos, making manual review unfeasible and calling for the use of artificial intelligence to detect relevant events in the footage \cite{schindler_action_2024}. 

Automatic action recognition (AR) and temporal action detection (AD) may be effective approaches for the identification of relevant events. Advanced techniques in the field of computer vision, especially pre-trained transformer \cite{fan_multiscale_2021, liu_video_2022, tong_videomae_2022} and foundation models (large vision models, LVMs) \cite{oquab_dinov2_2023}, have achieved high performance in AR and AD tasks. However, most of the studies in this field concern only the domain of human actions (with large benchmarking datasets such as Kinetics 400 \cite{kay_kinetics_2017}, Something Something \cite{goyal_something_2017}, and UCF101 \cite{soomro_ucf101_2012}).  In comparison, AR and AD in animal videos appear to be a much less researched field.

Existing papers on animal AR and AD tend to focus on a specific habitat or species, which are most often terrestrial large mammals \cite{schindler_action_2024, nishioka_detecting_2023, feng_action_2021, vogg_computer_2024}, while other types of animals (fish, birds, etc.) and habitats (water environments) are relatively neglected. Animal AR and AD studies most often apply custom datasets collected for a narrow application, which is understandable considering that the video analysis often serves the purpose of some specific animal study. On the other hand, this makes the results of these studies difficult to generalize over a wider domain. At the same time, there have been attempts to tackle AR and AD tasks on more diverse animal datasets (such as in \cite{mondal_actor-agnostic_2023}), most notably on Animal Kingdom \cite{ng_animal_2022}, which comprises videos of a variety of different species and environments. Such studies, however, are still relatively scarce compared to the abundance of human action recognition research.

Therefore, there is a scientific gap in approaches to identify actions (especially rare ones, such as predator attacks) in animal videos which can efficiently be applied to small and specific datasets (especially in scarcely researched domains, such as marine habitats), while also being able to generalize over larger and more diverse ones. This thesis addresses this gap by introducing the model MARINE (Motion-based Action Recognition In Natural Environments), which comprises motion-based frame selection in videos to identify the most relevant frames for fast animal actions, a pre-trained DINOv2 LVM backbone for feature extraction, and a lightweight trainable classification head to adapt to custom datasets.  To test its applicability to small and specific datasets, MARINE is applied in both an AR and AD task to a custom dataset of fish predator attacks at a coral reef, and to test its generalizability to more species and more varied environments, it is applied in an AR task to a subset of the Animal Kingdom dataset \cite{ng_animal_2022} comprising videos of predation-related actions of diverse fish species.

With this approach, the thesis aims to answer the following research question: \textit{To what extent can computer vision techniques be used to automatically detect rare ecological events, such as predation, in video footage of fish species?} This question can be divided into the following sub-questions:

\begin{enumerate}
    \item How can action recognition approaches be applied on a limited dataset of fish attack videos to classify and localize these rare events in the footage?
    \begin{enumerate}
        \item Specifically, can the model MARINE outperform VideoMAE, a state-of-the-art baseline in the AR and AD tasks on a custom coral reef dataset and a fish predation-related subset of Animal Kingdom?
    \end{enumerate}
    \item To what extent can intelligent frame selection improve the performance of recognition without the need to process the complete video, which is computationally more costly?
    \begin{enumerate}
        \item Specifically, does applying motion-based frame selection in the MARINE model instead of using the same number of evenly spaced frames increase the performance in the AR task?
    \end{enumerate}
    \item How can pre-trained vision foundation models be utilized within the action detection pipeline for feature extraction on video frames, so that only a small classification network needs to be trained on the custom dataset?
    \begin{enumerate}
        \item Does using a larger DINOv2 architecture in the MARINE model increase the performance on the AR task?
        \item Does using registers in the DINOv2 backbone, which are suggested to improve the quality of embeddings \cite{darcet_vision_2024}, increase MARINE's performance on the AR task?
        \item Does using a pre-trained foundation model ensure that MARINE performs well across animal domains other than fish predator attacks, specifically on a representative multi-label sample of Animal Kingdom?
    \end{enumerate}
\end{enumerate}

By answering these questions through a series of experiments with the MARINE model, the thesis aims to introduce an effective framework for tackling the automatic recognition of rare actions in animal videos, and thus contribute to the study of animals in their natural habitats.
\section{Related Work}
\label{sec:related_work}


The thesis is directed at filling the research gap on approaches in the automatic recognition of rare ecological events, specifically predator attacks among fish. Due to the relatively low amount of studies on this problem and the often narrow scope of the existing ones, it is expected that the project could offer a valuable contribution to the domain.

\subsection{Action recognition}
Action recognition (AR) is the task of classifying an action seen on video. Closely related is the task of action detection or temporal action localization (AD), which is the task of selecting frames depicting a certain action in a longer video sequence (i.e., finding the timestamps marking the beginning and end of an action within the video). In both cases, essential is the ability to encode and combine spatial and temporal information.

Earlier attempts at action recognition proposed 2D and 3D convolutional frameworks \cite{carreira_quo_2017, wang_spatiotemporal_2017}, which were designed to pool information from both the spatial and temporal dimension. As a notable example, the SlowFast family is known for applying two pathways, one at a lower frame rate to capture spatial information and one at high temporal resolution to capture fine motion \cite{feichtenhofer_slowfast_2019}. More recently, inspired by successes in the language processing domain, transformer-based models (vision transformers, ViT) made their way into the computer vision domain with excellent results \cite{dosovitskiy_image_2021}. Members of this family, such as MViT \cite{fan_multiscale_2021} and Swin \cite{liu_video_2022} consistently outperform their convolutional counterparts on common benchmark datasets. For instance, MViT and Swin both achieve over 81\% top-1 accuracy on Kinetics-400 and over 67\% on Something-Something v2, against Slowfast's 79.8\% and 63.1\%, respectively. What is more, the model VideoMAE utilizes an effective pre-training method using input masking with autoencoders, which results in state-of-the-art performance in both AR (75.4\% on Something-Something v2 and 87.4\% on Kinetics-400) and AD (39.3\% mean average precision, mAP, on AVA v2.2) tasks \cite{tong_videomae_2022}. It is important to note, however, that transformer models are computationally heavy with a vast amount of parameters (e.g., over 600 million for VideoMAE with the largest, ViT-H, backbone), which means that applying them to custom datasets usually involves fine-tuning some trainable parameters instead of the complete re-training of the model.

Furthermore, the emergence of foundation models, particularly Large Vision Models (LVM), is also an asset that can be capitalized upon in video understanding. These models are trained at scale on vast amounts of data, meaning that they may be used for diverse downstream tasks, potentially without the need of a task-specific labelled dataset (zero-shot learning) \cite{bommasani_opportunities_2021}. A notable example of the state-of-the-art in LVMs is DINOv2 \cite{oquab_dinov2_2023}, which is trained on a curated dataset created using self-supervised methods (a total of 142 million images), thus improving the quality of the extracted features, as opposed to using uncurated data. DINOv2's design builds on DINO \cite{caron_emerging_2021} and iBOT \cite{zhou_ibot_2022} and comprises a student and teacher network for both image-level and patch-level objectives (with separate loss terms for these two tasks). An important aspect of DINOv2's training process is a short high-resolution period at the end of training, which ensures that small objects do not disappear in the embedding. 

DINOv2, despite being trained solely on images, achieves comparable performance to the state-of-the-art in action recognition tasks through feature extraction on equally spaced frames (78.4\% accuracy on Kinetics-400) \cite{oquab_dinov2_2023}. In this approach, the embeddings produced by DINOv2 are passed through a simple linear classifier, which suggests that video understanding tasks may be solved without the need of training complex classifiers on video data, thus saving significantly on computational costs. What is more, a more recent addition to the DINOv2 architecture, register tokens, appear to enhance performance further by removing feature artifacts from non-informative background areas in an image \cite{darcet_vision_2024}, which might be particularly useful in recordings of natural environments, where the background may be dense and noisy.

Efficiency is at the heart of many studies in the domain of video understanding, as these tasks generally require large computational resources. One important implication of this is that it is rarely possible for a model to consider every frame in a video: it is much more common to select only some of them. To resolve this problem, Liu et al.  \cite{liu_no_2021} propose a full-video action recognition approach which efficiently utilizes information from every frame through dynamic temporal clustering.  At the same time, informed frame selection may also be an effective and computationally efficient method to identify the most relevant frames for a video understanding task. Notably, Gowda et al. \cite{gowda_smart_2021} propose an intelligent frame selection approach, SMART, which outperforms other frame selection approaches (such as AdaFrame \cite{wu_adaframe_2019} and FrameGlimpse \cite{yeung_end--end_2016}) and even full video action recognition on the UCF-101 dataset. The reasoning behind the latter is that it discards frames which are difficult to classify and thus confuse the decision. However, the implementation of these frame selection techniques is either not published (\cite{wu_adaframe_2019, gowda_smart_2021}), or comprises more of an end-to-end model which can be difficult to integrate into a custom AR or AD pipeline such as MARINE, necessitating the construction of a lightweight custom frame selection method inspired by these techniques.

\subsection{Animal action recognition and detection}
\label{sec:animal_AR}
Research on action recognition in the animal domain is still sparse compared to human actions. What is more, existing studies focus mainly on a narrow range of (or even a single) species, which are most often large mammals. In \cite{schindler_action_2024}, Schindler et al. introduce MAROON, an action recognition system relying on segmentation and deployed on a self-assembled dataset of deer camera trap footage. In \cite{nishioka_detecting_2023}, Nishioka et al. use temporal action localization to monitor elephant eating behavior in a zoo, while in \cite{feng_action_2021}, Feng et al. focus specifically on feline action recognition in the wild. These studies rely heavily on the identification of the actor, an aspect which is also highlighted in \cite{vogg_computer_2024}, a study on current and future directions in computer vision for primate behavior studies. In such approaches, first the (potential) actor is identified in the footage, then action classification follows based primarily on the actor's visual features \cite{schindler_action_2024, nishioka_detecting_2023} or estimated pose \cite{feng_action_2021}.

This type of approach, however, carries disadvantages in settings when a single actor is difficult to identify, or when its appearance or pose cannot easily be discerned. Schools of fish in underwater video footage are such an example, and this might partly be the reason why video understanding studies on fish videos are especially rare. Existing research mainly concerns aquacultures, such as \cite{maloy_spatio-temporal_2019}, in which the combination of a 3D convolutional network and Long Short-Term Memory is used to classify whether salmon are feeding or not. While this model achieves good performance on the (small) dataset used in the study (80.0\% average accuracy), it is likely that more advanced methods such as vision transformers could achieve higher results and generalize better over other data samples.

MSQNet is a welcome example of an action recognition model which achieves high performance on a diverse animal dataset, Animal Kingdom \cite{ng_animal_2022}, and which does not rely on the identification of the actor \cite{mondal_actor-agnostic_2023}. Due to the CLIP-based image-language embedder included in its architecture, MSQNet is capable of utilizing textual cues, thus possessing zero-shot capabilities. As such, it offers a viable solution for tackling video understanding tasks even in animal footage which depicts previously unseen species. On Animal Kingdom, MSQNet achieves 73.10\% mAP \cite{mondal_actor-agnostic_2023}, which is significantly higher than I3D's 16.48\%, Slowfast's 20.45\%, or X3D's 25.25\% (as reported in \cite{ng_animal_2022}).

\subsection{Animal video datasets}
\label{sec:related_work_animalDatasets}
Although ecological and ethological studies could greatly benefit from computational video understanding through the automation of tasks listed in and similar to Section \ref{sec:animal_AR}, there is a relative scarcity in publicly available animal video datasets. Many of the studies in this area utilize only a small and specific dataset which is often compiled particularly for the study's purpose (e.g., \cite{schindler_action_2024, nishioka_detecting_2023, maloy_spatio-temporal_2019, feng_action_2021} of the ones described in Section \ref{sec:animal_AR}). Besides focusing on a specific environment or species, these datasets are often unpublished, making it difficult for future studies to build on.

Of the existing public datasets, a majority target image-based tasks or animal recognition in videos, less so the understanding of temporal dynamics. Good examples are iNaturalist \cite{van_horn_inaturalist_2018} and Animals with Attributes \cite{lampert_attribute-based_2014}, both of which encompass a wide range of species and conditions, but comprise only images. Therefore, if used in video understanding tasks, the temporal dimension would have to be neglected.

In \cite{ng_animal_2022}, Ng et al. introduce Animal Kingdom, a comprehensive animal video dataset collected from YouTube and complete with annotations for several different tasks, among them action recognition. The true advantage of this dataset is not only its relatively large size (over 30,000 clips in case of action recognition), but also its diversity. The videos depict a total of 850 species from six large animal groups in varying environments (underwater, desert, forest, etc.) and conditions (snow, night vision, etc.). The annotations include 140 different action classes. Such a varied dataset can be an important asset in building models for action classification tasks in the animal domain. At the same time, as explained in \cite{ng_animal_2022}, broad datasets such as Animal Kingdom tend to possess a long-tailed distribution of annotated action classes, as rare animal behaviors are inherently less documented. It is yet to be explored how well such a diverse dataset can be adopted in ecological studies focusing only on a single or few animal action(s), species, or habitat(s), especially if these belong to minority classes.

In summary, video understanding is a dynamically developing field within computer vision, but most of the research in the AR and AD area has been dedicated to human actions, with animal videos being neglected in comparison. Existing research in the animal domain predominantly focuses on actor-based action recognition and often relies on rather outdated methods and small custom datasets. What is more, water habitats and non-mammal species are generally underrepresented in animal video research. This thesis aims to address this gap by introducing a methodology to utilize state-of-the-art computer vision techniques in AR and AD tasks to identify fish predator attacks, using both a small and specific dataset and a subset of a larger and more diverse animal benchmark dataset.
\section{Methodology}
\label{sec:methodology}


The methodology concerns the construction of the MARINE model, comprising a motion-based frame selection module, DINOv2 feature extraction, and a trainable classification head, and its evaluation in a binary AR (classifying whether a predator attack occurs in a short video clip) and AD (temporally localizing predator attacks in longer recordings) task on two datasets of fish videos.

\subsection{Datasets}
\label{sec:method_data}

The study relies mainly on two datasets: Animal Kingdom (AK), as presented by Ng et al. \cite{ng_animal_2022}, and a small custom dataset referred to as the coral reef dataset. These datasets are presented in more detail in Sections \ref{sec:method_data_AK} and \ref{sec:method_data_coralReef}, respectively.

\begin{figure*}[!t]
  \centering
  \includegraphics[width=\textwidth]{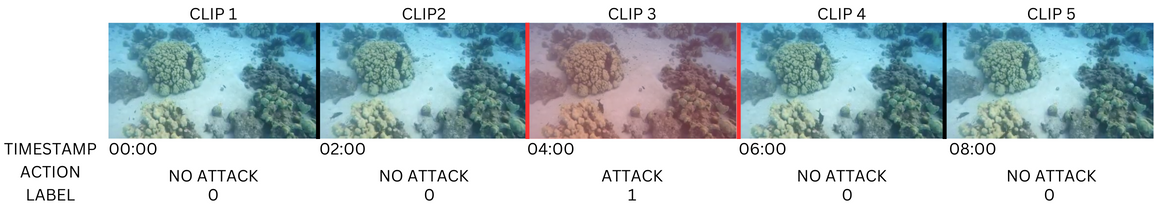}
  \caption{Processing the original coral reef dataset. Each of the 44 10-second-long videos are separated into 5 clips, with the middle one containing a predator attack. The resulting 2-second-long clips (220 in total) will comprise the processed coral reef dataset, where the clips originally situated at the centre of the uncut videos receive the positive label and all others the negative label.}
  \label{fig:fish_video_attack_pos}
\end{figure*}

\subsubsection{Animal Kingdom}
\label{sec:method_data_AK}
Animal Kingdom (AK) is a dataset annotated for several different video understanding tasks, of which the action recognition dataset, comprising over 850 animal species and 140 fine-grained action classes \cite{ng_animal_2022}, is utilized in this study. As the thesis focuses on predator attacks observed in underwater habitats, the AK action recognition dataset is filtered for videos which depict fish. To avoid confusion, this narrower dataset will be referred to as the AK fish dataset throughout the paper.

The AK fish dataset is filtered from the AK action recognition dataset by selecting those videos in which a fish species performs some (any) action. Out of these, videos containing predator attacks are specified as ones in which one of the following eight predation-related actions, taken from the fine-grained action labels of AK, is performed by a fish species (for an exact description of these actions, see Appendix \ref{sec:apx:AK_fish_description}):

\begin{multicols}{2}
\begin{itemize}
    \item Attacking
    \item Being eaten
    \item Biting
    \item Chasing
    \item Fighting
    \item Fleeing
    \item Retaliating
    \item Struggling
\end{itemize}
\end{multicols}

There are 887 video clips in the AK fish dataset, out of which 81 contain an attack-related action performed by fish.  Table \ref{tab:datasets} shows a summary of the dataset, while a more comprehensive description is shown in Appendix \ref{sec:apx:AK_fish_description}. 

For the purpose of the AR task (i.e., identifying whether a video contains an attack or not), the videos in the AK fish dataset receive binary labels: positive if they contain one of the predation-related actions performed by a fish species, and negative otherwise. The resulting dataset suffers from high class imbalance, where only $9.13\%$ of videos are labelled as positive. This factor must be considered during model training, as explained in Section \ref{sec:method_experiments_AR}.

\subsubsection{Coral reef dataset}
\label{sec:method_data_coralReef}
The coral reef dataset, recorded by members of the Institute of Biodiversity and Ecosystem Dynamics at the University of Amsterdam, comprises 44 videos, each of which are 10 seconds long. These videos were recorded in high spatial and temporal resolution (1080x1920 pixels and 120 frames per second). Each of the videos show a similar coral reef environment and contain a fish predator attack in the central section of the video (within a 2-second time window around the middle point), as illustrated in Figure \ref{fig:fish_video_attack_pos}.

For the purpose of the binary action recognition task, the coral reef dataset is processed through the following steps, which are also explained in Figure \ref{fig:fish_video_attack_pos}:

\begin{enumerate}
    \item The videos are separated into 2-second-long clips without overlaps, resulting in 5 clips cut from each 10-second long video.
    \item As a predator attack is contained in the middle section of each video, the resulting clips are labelled so that the centre clip is labelled as positive, and the four other clips are labelled as negative.
\end{enumerate}

The resulting dataset comprises 220 videos, of which 44 (the centre clips of the original 44 videos) contain an attack and receive a positive label. When mentioning the coral reef dataset from here onwards in the paper, this processed dataset is referred. The coral reef dataset, summarized in Table \ref{tab:datasets},  also suffers from class imbalance, with only $20\%$ of instances comprising the positive class.

\begin{table}[]
\small
\centering
\begin{tabular}{lll}
                                & AK fish & Coral reef \\ \hline
Number of videos                & 887     & 220        \\
Positive ("attack") class (\% of sample) & 9.13    & 20.00         \\
Number of frames (mean)         & 117.68  & 239.55     \\
Number of frames (st. dev.)     & 92.71   & 0.61       \\ \hline
\end{tabular}
\caption{Summary table of the AK fish and coral reef datasets. Both datasets comprise binary classes ("attack" or "no attack"). }
\label{tab:datasets}
\end{table}

\subsection{Action recognition model}
\label{sec:method_AR}

\subsubsection{General framework}
\label{sec:method_AR_general}

The action recognition task of identifying whether a video contains a predator attack can be understood as a binary classification problem. In general terms, the model proposed for this task, MARINE, comprises the following components (also referred to as modules), which are illustrated in Figure \ref{fig:pipeline}:

\begin{enumerate}
    \item \textbf{Frame selection module.} To increase computational efficiency, not every frame from the video is considered for action recognition. Instead, only a limited number of frames are selected and processed for classification.

    \item \textbf{Image embedding using DINOv2.} The image foundation model DINOv2 \cite{oquab_dinov2_2023} is utilized for feature extraction on the selected frames. The resulting frame features are concatenated to retain information from the temporal dimension of the video.

    \item \textbf{Shallow classification head.} Using the concatenated features resulting from the previous step as input, a relatively simple classifier network is trained to distinguish between videos which do or do not contain a predator attack.
    
\end{enumerate}

The novelty and contribution of this general approach is three-fold. Firstly, it integrates two important techniques in action recognition: intelligent frame selection methods, which enable classification models to focus on the most relevant frames of a video, and large vision models, which provide a robust solution for extracting meaningful features from images without the need for extensive training. Secondly, it tackles videos in the domain of animal actions, which are considerably less researched than videos depicting human actions. Lastly, it promotes computational efficiency in several of its aspects:

\begin{itemize}
    \item Allowing to process only a limited number of frames from the video, which requires considerably less resources than processing the entire clip, especially if it is high resolution, as the coral reef dataset is.

    \item Utilizing a pre-trained image foundation model without additional training. DINOv2 has been shown to achieve state-of-the-art performance in action recognition tasks without being fine-tuned on the video dataset \cite{oquab_dinov2_2023}.

    \item Using a shallow classifier network at the head of the action recognition pipeline. Using the extracted features as input, a small classifier is trained instead of a computationally heavy deep learning network.
\end{itemize}

\subsubsection{Module specifics}
\label{sec:method_AR_modules}
The details of MARINE's modules are as follows.

\begin{itemize}
    
    \item \textbf{Frame selection module.} Predator attacks are characterized by swift movements, which can be especially pronounced visually if the attack happens in a school of fish. Therefore, in the proposed frame selection method, those frames are selected for inspection in which rapid movement can be detected. This is done by assessing the dissimilarity between two consecutive frames and selecting a set number of frames (specifically 10) which are most dissimilar to the one before them. The dissimilarity score $\alpha_{p,q}$ can be calculated between two frames represented as tensors $W^p$ and $W^q$, each with $n$ elements, in the following way:

    \begin{equation}
    \label{eq:dissimilarity}
        \alpha_{p,q} = \sum_i^n |W_i^p - W_i^q|
    \end{equation}

As changes in individual color channels may not indicate motion, frames are converted to greyscale images and represented as two-dimensional matrices during motion-based frame selection.

    \item \textbf{Feature extraction using DINOv2.} Image embeddings of the selected frames are obtained by retrieving the classifier tokens from the DINOv2 backbone's output, without fine-tuning the model to maximize computational efficiency. The features are then concatenated into a single vector to retain temporal information from the video. Four different DINOv2 backbones are tested as part of the MARINE model (as described in more detail in Section \ref{sec:method_experiments_AR}).

    \item \textbf{Classification model.} After obtaining the concatenated frame features, the resulting video-level vectors are used as input to a classifier head, a shallow neural network with a final sigmoid activation. The exact settings of the network are determined through cross-validation, as described in Section \ref{sec:method_experiments_AR}.
        
\end{itemize}

\subsection{Action detection in the coral reef dataset}
\label{sec:method_AD}

\begin{figure*}[!t]
  \centering
  \includegraphics[width=\textwidth]{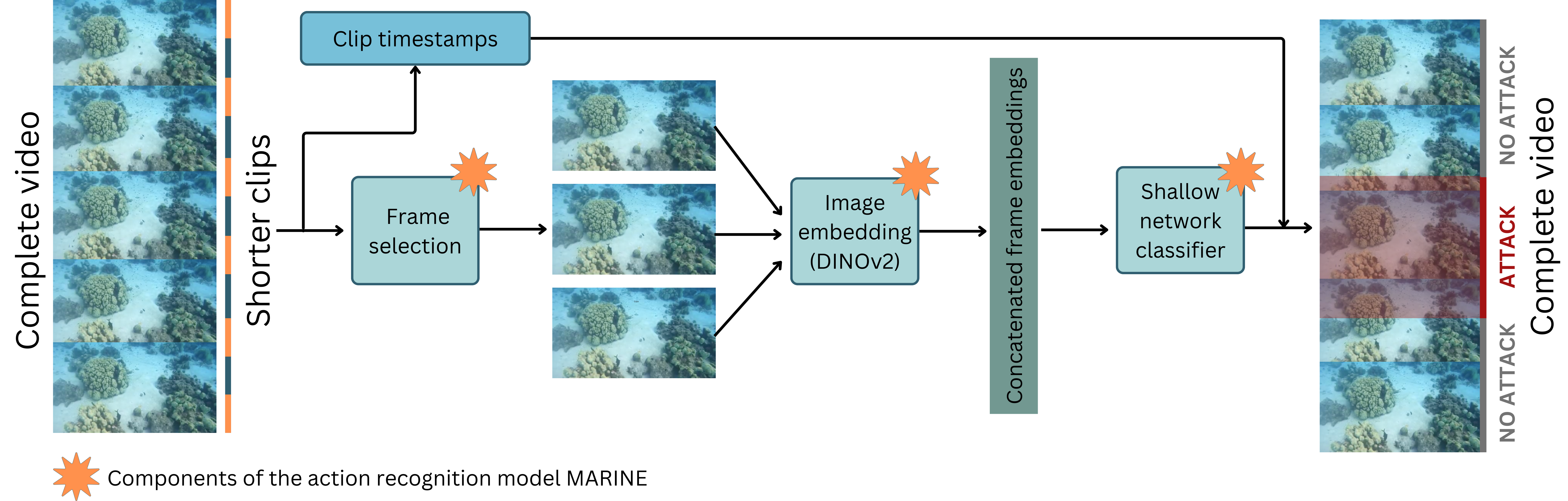}
  \caption{Action detection pipeline. The untrimmed video is separated into shorter clips, which are passed to the action recognition model. Combining the predictions of this model and the timestamps which locate the clips in the original video, the temporal location of the predator attack can be predicted in the original footage.}
  \label{fig:pipeline}
\end{figure*}

Besides being able to classify a single short clip as containing a predator attack or not, a method to localize attacks in longer untrimmed videos (AD task) is proposed. This approach builds on the AR model MARINE and consists of the following steps, also summarized in Figure \ref{fig:pipeline}:

\begin{enumerate}
    \item Separate the untrimmed video into shorter clips and store the delimiting timestamps of each clip in the original video. These separated clips coincide with the ones described in Section \ref{sec:method_data_coralReef} and Figure \ref{fig:fish_video_attack_pos}, i.e., the processed coral reef dataset.

    \item Classify each of the short clips as containing a predator attack or not using MARINE.

    \item Using the predicted label of each short clip, as well as the timestamps preserved when cutting the clips from the untrimmed footage, identify the sections in the original video which are classified as containing an attack.
    
\end{enumerate}

\subsection{Evaluation}
\label{sec:method_eval}

\subsubsection{Action recognition}
For evaluating MARINE's performance on the AR task (both in the AK fish and coral reef datasets), the commonly used metrics accuracy, precision, recall, and F1-score are applied. The latter three are especially important to consider due to the high class imbalance observed in the data. ROC curves will further help clarify the models' ability to balance between recall and maintaining a low false positive rate.

\subsubsection{Action detection in the coral reef dataset}
To assess the performance of temporal action detection, mean average precision (mAP) is most commonly used \cite{xia_survey_2020}. This metric is suitable for multi-label classification - in a binary case as in this study, average precision (AP, the mean precision over all videos) suffices. To calculate precision, the correctness of each predicted action time interval must be decided. Following \cite{xia_survey_2020}, t-IoU (temporal Intersection over Union, as shown in Equation \ref{eq:t-iou}) with a threshold of 50\% is applied for this purpose.

\begin{equation}
\label{eq:t-iou}
    \textit{t-IoU} = \frac{\textit{predicted time interval} \cap \textit{true time interval}}{\textit{predicted time interval} \cup \textit{true time interval}}
\end{equation}

Note that a post-hoc test with a t-IoU threshold of 25\% is also conducted, please see details in Section \ref{sec:results_AD}.

\subsubsection{Existing benchmarks}
While extensive benchmarks exist for human action recognition problems, it is less so the case for animal datasets. On human action datasets in both AR and AD tasks (on the Kinetics 400 and AVA v2.2 datasets), VideoMAE achieves state-of-the-art performance, obtaining the highest results of all models reviewed in Section \ref{sec:related_work}. As VideoMAE also demonstrates remarkable transfer learning capabilites \cite{tong_videomae_2022}, it was chosen to serve as the baseline model to apply to the AK fish and coral reef datasets to benchmark MARINE's performance. The specifics of fine-tuning VideoMAE for the study are presented in Section \ref{sec:method_experiments_AR}.

At the same time, limited benchmarks do exists on animal video datasets, specifically for Animal Kingdom, including the models MSQNet, I3D, Slowfast and X3D, as discussed in Section \ref{sec:related_work_animalDatasets}. As an additional test of its applicability across animal domains, MARINE is adapted to a multi-label setting and compared to these benchmarks on a representative subsample of Animal Kingdom (as is explained in Section \ref{sec:method_experiments_AK-sample}).

\subsection{Experimental setup}
\label{sec:method_experiments}

Informed by the research questions introduced in Section \ref{sec:introduction} and the objectives of evaluation detailed in Section \ref{sec:method_eval}, the study applies the experimental setup detailed below. Implementation of all models and experiments can be found on the project's GitHub page.

\subsubsection{Action recognition.}
\label{sec:method_experiments_AR}
To address research questions (1) and (3), the performance on the action recognition task is measured on two datasets: the AK fish dataset and the coral reef dataset. As both of these datasets can be considered quite small in the domain of video understanding, a bootstrapped test sample is used, so that results can be reported with confidence intervals. This is done by defining training and test samples within both datasets: as Animal Kingdom is published with pre-defined train and test sets, the same split is kept within the AK fish sample, while in the coral reef dataset, a random split with 20\% test size is performed (ensuring that clips separated from the same video, as shown in Figure \ref{fig:fish_video_attack_pos}, are assigned to the same split). After training the different MARINE setups (and fine-tuning VideoMAE), all models are evaluated on bootstrapped test sets of the two datasets: the test set is resampled with replacement 100 times (with the number of samples remaining identical to the original test set), and the mean accuracy, precision, recall, and F1-score are reported with standard deviation in the bootstrapped test set. 

The training process of MARINE starts with using the motion-based frame selection method to identify the 10 most relevant frames, then applying DINOv2 feature extraction to the selected frames (as described in \ref{sec:method_AR_modules}). The experiments use four different DINOv2 backbones: ViT-S/14 (MARINE S14), comprising 21 million parameters, and ViT-G/14 (MARINE G14), comprising 1,100 million parameters, both architectures with and without register tokens. Testing different backbones specifically addresses sub-questions (a) and (b) of reseach question (3). In MARINE, feature extraction refers to obtaining the classifier token of the DINOv2 output when the frame is passed to the model.

After concatenating the extracted frame features, the classifier head is trained using 3-fold cross-validation. This classifier is a shallow neural network with a 10-node dense input layer with ReLu activation, between 0 and 3 dense hidden layers with 128 ReLu-activated nodes each, and a final sigmoid output node. Importantly, class weights are utilized during training to address the imbalance in both datasets. The threshold to consider the sigmoid output of the network as predicted positive is determined post-training by optimizing for F1-score. The specifics of the training process, including the hyperparameter settings of the classifier, are presented in Appendix \ref{sec:apx:training}.

Regarding the benchmark model, the classification head of a VideoMAE model with ViT-B backbone pretrained on Kinetics 400 is fine-tuned on the training sets of the AK fish and coral reef datasets, using video frames selected through the same motion-based method as in MARINE. The threshold to consider the continuous output of the model as predicted positive is also determined identically to MARINE, by optimizing for F1-score. The implementation of the fine-tuning process is detailed in Appendix \ref{sec:apx:VideoMAE}.

\subsubsection{Ablation study for frame selection module.}
To address research question (2), the contribution of the motion-based frame selection module in MARINE is assessed in an ablation study setting. In this experiment, the performance of the MARINE G14 model is compared to a MARINE G14 model where the motion-based frame selection module is swapped for one which selects 10 evenly spaced frames, starting from the first one in the video. Performance is measured on the same bootstrapped test sets as described in Section \ref{sec:method_experiments_AR}, reporting the same metrics (accuracy, recall, precision, and F1-score).

\subsubsection{Action detection.}
\label{sec:method_experiments_AD}
Performance in the action detection task is measured through AP relying on t-IoU, as formulated in Equation \ref{eq:t-iou}. As localized action annotation is available only for the coral reef dataset, the models (MARINE setups and VideoMAE) will be evaluated on this dataset for this task. Generating the predicted temporal location of attacks is done as described in Section \ref{sec:method_AD} and Figure \ref{fig:pipeline}, and the difference in the experimental runs lies in the selected AR model: different MARINE setups and VideoMAE, trained/fine-tuned using the process described in Section \ref{sec:method_experiments_AR}. Importantly, the test set for this task comprises only 9 videos (20\% of the 44 untrimmed videos in the coral reef dataset), and to report results, a bootstrapped test set is once more used. This means that a sample of 9 videos is selected from the test set, with replacement, 100 times, and the mean of the average precisions on these 100 samples is reported, as well as its standard deviation. The size of the individual test samples is kept identical to the original test set, as small sample sizes are characteristic to real-world biological datasets. At the same time, resampling 100 times helps introduce variation, so that the results may be reported with statistical confidence.

\subsubsection{Multi-label benchmarking on Animal Kingdom.}
\label{sec:method_experiments_AK-sample}
To address sub-question (c) of research question (3), namely whether MARINE can perform well across varied animal domains, its performance is tested on a representative sample of Animal Kingdom in a multi-label setting. This random sample comprises 1000 videos and 93 classes of the Animal Kingdom action recognition dataset, and its distribution of labels is confirmed to be representative of the original dataset at 95\% confidence with a chi-squared test of homogeneity (p-value = 0.68). MARINE G14 (without registers) is adapted to this multi-label setting by replacing the single sigmoid output of the classification head with a layer of 93 sigmoid nodes.

Using the pre-defined train and test split of the Animal Kingdom sample, motion-based frame selection and DINOv2 feature extraction is performed on both splits, while the multi-label classification head is trained using cross validation on the training set (similarly to the binary classification setting described in \ref{sec:method_experiments_AR}). Finally, predictions are obtained for the test set and mAP is calculated as the main performance metric to compare with existing benchmarks published in \cite{mondal_actor-agnostic_2023} and \cite{ng_animal_2022}. All specifics of this experimental setup can be found in Appendix \ref{sec:apx:AK-sample}.

\section{Results}
\label{sec:results}


\subsection{Action recognition results}
\label{sec:results_AR}

To address research questions (1) and (3), specifically whether MARINE can outperform VideoMAE in the AR task and how different DINOv2 backbones influence its performance, Table \ref{tab:AR_results} summarizes the outcomes of the experiments described in Section \ref{sec:method_experiments_AR}. For clarity's sake, and as the presence of register tokens in the DINOv2 backbones does not result in any obvious performance increase, Table \ref{tab:AR_results} only reports the performance of the MARINE S14 and G14 models without registers.  Appendix \ref{sec:apx:registers} shows the outcomes of experiments run using MARINE models containing DINOv2 backbones with registers.

Table \ref{tab:AR_results} shows the performance of the proposed models in terms of accuracy, recall, precision, and F1-score (mean $\pm$ 1.96 standard deviations to indicate the 95\% confidence interval) on the binary classification task of identifying videos which contain a predator attack. The results indicate that MARINE G14 is capable of better performance than VideoMAE in terms of accuracy, precision and F1-score on both datasets, and in terms of all four metrics on the AK fish dataset. On this dataset, MARINE G14 achieves over 10\% points increase in accuracy compared to the other two models, and over 30\% points in F1-score.

Notably, the performance of all three models is considerably higher on the AK fish dataset. Achieving a good balance between precision and recall appears especially problematic on the coral reef dataset, where MARINE S14 achieves 100.0\% recall with 0.00\% points variation, but low precision (19.51\%). The fact that the accuracy in this case (19.51\%) is very close to the proportion of positive samples in the dataset (20.0\%) indicates that the model essentially classifies all instances as positive. At the same time, MARINE G14 appears more capable of balancing precision and recall although at rather low values (52.52\% and 53.01\%, respectively).

While these results suggest that MARINE G14 could be capable of state-of-the-art performance on relatively small animal video datasets of specific animal types and environments, it must be noted that the limited test sample size results in large variations in the outcomes, as indicated by the wide confidence intervals.

\begin{table*}[]
\begin{tabular}{llllll}
Dataset     & Model      & Accuracy (\%)           & Recall (\%)             & Precision (\%)          & F1-score (\%)           \\ \hline
Coral reef& MARINE S14 & 19.51 +/ 12.25          & \textbf{100.00 +/ 0.00} & 19.51 +/ 12.25          & 32.19 +/ 17.30          \\
Coral reef& MARINE G14 & \textbf{81.53 +/ 10.36} & 52.52 +/ 33.77          & \textbf{53.01 +/ 37.24} & \textbf{51.06 +/ 29.60} \\
Coral reef& VideoMAE   & 52.64 +/ 13.92& 69.72 +/ 31.33 & 24.51 +/ 16.77& 35.49 +/ 20.33\\ \hline
AK fish     & MARINE S14 & 81.74 +/ 7.05           & 57.40 +/ 41.50          & 16.07 +/ 14.68          & 24.62 +/ 20.54          \\
AK fish     & MARINE G14 & \textbf{94.86 +/ 3.97}  & \textbf{63.31 +/ 41.60} & \textbf{51.64 +/ 36.86} & \textbf{55.14 +/ 33.18} \\
AK fish     & VideoMAE   & 83.14 +/ 7.70           & 42.73 +/ 43.08          & 14.33 +/ 16.76          & 20.92 +/ 22.76         
\end{tabular}
\caption{Model performances on the action recognition task, using motion-based frame selection with each model and DINOv2 backbones without registers in the MARINE models. The table shows the performance on the AR task in terms of accuracy, recall and precision ($\pm$ 1.96 standard deviations to indicate the 95\% confidence intervals).}
\label{tab:AR_results}
\end{table*}

\begin{figure}
    \centering
    \includegraphics[width=0.9\linewidth]{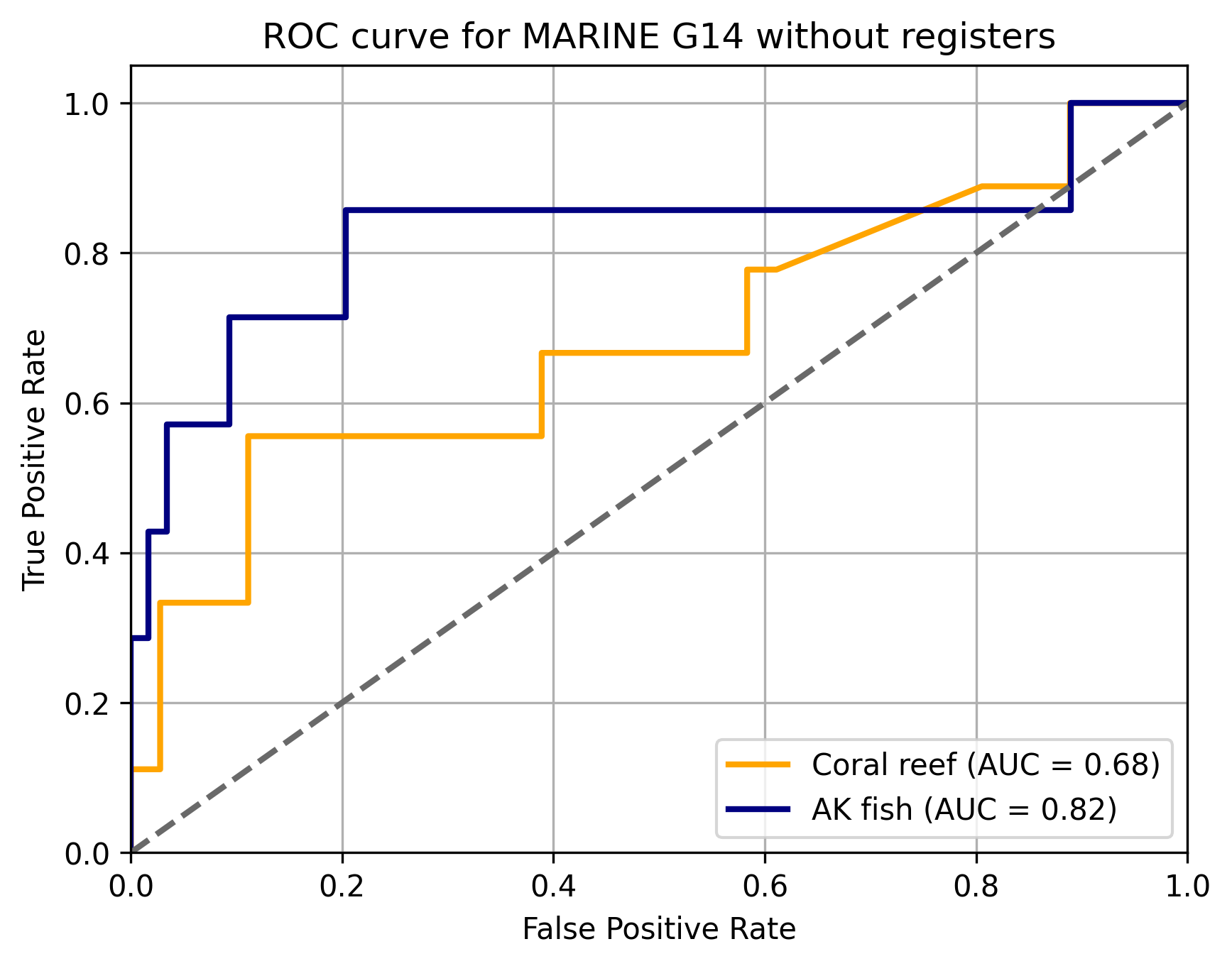}
    \caption{ROC curves of MARINE G14 on the coral reef and AK fish datasets, without registers in the DINOv2 backbone. The higher AUC value for the AK fish dataset is consistent across all configurations of the MARINE model.}
    \label{fig:ROC}
\end{figure}

The challenging nature of the coral reef dataset is highlighted by Figure \ref{fig:ROC}, which shows the ROC curves of MARINE G14 on both test sets. The area-under-the-curve (AUC) is a relevant measure in the AR task as the target action, a predator attack, is extremely rare, which results in highly imbalanced datasets. Therefore, it is essential that the model is able to find an appropriate balance between identifying all relevant instances (true positive rate) without extensively compromising on specificity (or, conversely, on the false positive rate). Figure \ref{fig:ROC} clearly shows that MARINE G14 was more capable of this on the AK fish dataset than on the coral reef dataset (AUC of 0.82 and 0.68, respectively). The fact that this outcome is consistent across all configurations of the MARINE model indicates that achieving an appropriate recall-false positive rate balance is generally more challenging in case of the coral reef dataset.

\subsection{Ablation study for motion-based frame selection}
An ablation study was conducted to address research question (2) and assess the contribution of the motion-based frame selection method in MARINE. In this experiment, MARINE S14 and G14 were tested against their counterparts using evenly selected frames instead of the motion-based frame selection method. The results are presented in Table \ref{tab:ablation_results} (without the use of register tokens). As the table shows, applying motion-based frame selection does not clearly contribute to higher performance. Specifically, accuracy decreases on the AK fish dataset but increases significantly (from 23.71\% to 81.53\%) on the coral reef dataset when motion-based frame selection is introduced in MARINE G14. At the same time, while motion-based frame selection does not consistently improve F1-scores, it appears to help avoid an extreme imbalance between precision and recall, where 100\% performance is achieved in one of the metrics and (extremely) low performance on the other, indicating that the model either classifies virtually all instances as positive (100\% recall, low precision) or as negative (100\% precision, low recall). Notably, however, this behaviour occurs even using motion-based frame selection with MARINE S14 on the coral reef dataset, which indicates that such a challenging dataset requires not only more sophisticated frame selection but also more complex feature extraction using larger LVM backbones.

\begin{table*}[]
\begin{tabular}{lllllll}
Dataset    & Frame selection & Model      & Accuracy (\%)           & Recall (\%)             & Precision (\%)          & F1-score (\%)           \\ \hline
Coral reef & Evenly spaced   & MARINE S14 & 19.51 +/ 12.25          & 100.00 +/ 0.00          & 19.51 +/ 12.25          & 32.19 +/ 17.30          \\
Coral reef & Motion-based    & MARINE S14 & 19.51 +/ 12.25          & 100.00 +/ 0.00          & 19.51 +/ 12.25          & 32.19 +/ 17.30          \\ \hline
Coral reef & Evenly spaced   & MARINE G14 & 23.71 +/ 12.53          & \textbf{100.00 +/ 0.00} & 20.35 +/ 12.70          & 33.34 +/ 17.66          \\
Coral reef & Motion-based    & MARINE G14 & \textbf{81.53 +/ 10.36} & 52.52 +/ 33.77          & \textbf{53.01 +/ 37.24} & \textbf{51.06 +/ 29.60} \\ \hline
AK fish    & Evenly spaced   & MARINE S14 & \textbf{96.65 +/ 3.18}  & 45.99 +/ 32.53          & \textbf{100.00 +/ 0.00} & \textbf{61.27 +/ 30.12} \\
AK fish    & Motion-based    & MARINE S14 & 81.74 +/ 7.05           & \textbf{57.40 +/ 41.50} & 16.07 +/ 14.68          & 24.62 +/ 20.54          \\ \hline
AK fish    & Evenly spaced   & MARINE G14 & \textbf{97.12 +/ 3.01}  & 52.63 +/ 36.55          & \textbf{100.00 +/ 0.00} & \textbf{67.07 +/ 30.66} \\
AK fish    & Motion-based    & MARINE G14 & 94.86 +/ 3.97           & \textbf{63.31 +/ 41.60} & 51.64 +/ 36.86          & 55.14 +/ 33.18          \\ \hline
\end{tabular}
\caption{Results of the ablation experiment to test the contribution of the motion-based frame selection method to the AR task using the MARINE models without registers in the DINOv2 backbones. The highest performances in terms of each metric on both datasets are highlighted in bold.}
\label{tab:ablation_results}
\end{table*}

\subsection{Action detection results}
\label{sec:results_AD}
To address research question (1), namely how the proposed MARINE model performs in the AD task in comparison to VideoMAE, Table \ref{tab:AD_results} presents the results of this task on the coral reef dataset in terms of average precision (mean over bootstrapped test samples and standard deviation). The presented results were obtained applying motion-based frame selection in all models, and applying DINOv2 backbones without registers in the MARINE models. 

The most important outcome of the AD task experiments is that applying a threshold of 50\% t-IoU for considering a predicted time interval as correct yielded no correct predictions for any of the models (MARINE or VideoMAE), which, similarly to outcomes reported in Section \ref{sec:results_AR}, highlights the challenging nature of the coral reef dataset. Therefore, it is crucial to understand that the results reported in Table \ref{tab:AD_results} were obtained using a lower t-IoU threshold of 25\%, meaning that a predicted time interval is considered correct if it coincides with a true attack in at least 25\% of its duration.

Using the lowered threshold, MARINE G14 outperforms both other models by a substantial margin (80.78\% against VideoMAE's 34.89\% and MARINE S14's 0.00\%). At the same time, the non-varyingly low performance of MARINE S14 (0.00\% with a standard deviation of 0.00\%) shows not only this model's poor ability to temporally localize predator attacks in the coral reef dataset, but also the limitations introduced by the small size of the test sample. The fact that MARINE S14's performance is non-varyingly 0.00\% indicates that it cannot correctly predict a single attack timeframe at a threshold of 25\% t-IoU in any of the 9 test videos from which the bootstrapped samples are taken (as explained in Section \ref{sec:method_experiments_AD}).

\begin{table}[]
\begin{tabular}{lll}
Model      & Average precision& Standard deviation \\ \hline
MARINE S14 & 0.00\%            & 0.00\%             \\
MARINE G14 & \textbf{80.78\%}  & 12.05\%            \\
VideoMAE   & 34.89\%& 14.32\%\\ \hline
\end{tabular}
\caption{Model performances on the action detection task on the coral reef dataset, in terms of average precision (AP) at a lowered t-IoU threshold of 25\%. The values relate to the mean and standard deviation of average precisions in the bootstrapped samples of test videos. Motion-based frame selection is applied with all models, and DINOv2 backbones do not apply registers. The highest performance is highlighted in bold.}
\label{tab:AD_results}
\end{table}

\vspace{-0.3\baselineskip}
\subsection{Multi-label classification results}
\label{sec:results_AK-sample}
To address sub-question (c) of research question (3) and investigate the extent to which MARINE may generalize across varied animal domains, its performance was tested in a multi-label setting on a representative sample of Animal Kingdom comprising 93 classes. In this experiment, MARINE G14 (without registers) achieves 23.79\% mAP, which is significantly lower than MSQNet's 73.10\% \cite{mondal_actor-agnostic_2023}, comparable to X3D's 25.25\%, and higher than Slowfast's 20.45\% and I3D's 16.48\% \cite{ng_animal_2022}. This shows that although MARINE could be a viable option in a binary classification setting, it cannot currently compete with the best of the state-of-the-art on a diverse multi-label animal dataset.

\section{Discussion}
\label{sec:discussion}

\subsection{Most important outcomes}
The aim of the study is to introduce a computer vision model, MARINE, which can be applied to tackle AR and AD tasks in animal video datasets, specifically to identify fish predator attacks. In terms of the AR task, the results in Section \ref{sec:results} demonstrate that MARINE performs comparably or better in the AR task than state-of-the-art vision transformer model VideoMAE with a ViT-B backbone. Comparing the results to other animal action recognition studies which apply datasets similarly specific to some habitat and/or species, Table \ref{tab:comparison_studies} shows that MARINE G14 performs comparably to all referenced models and outperforms several of them on its own specific domain and dataset. Importantly, each of the results shown in Table \ref{tab:comparison_studies} was obtained on different animal datasets used in the specific studies. The similarity between the datasets is that each comprised videos of a highly specific habitat and species, as described in the table. These results indicate that MARINE is a viable option for AR tasks on animal video samples related to some specific domain (predator attacks among fish in this case), but for a more conclusive comparison, each of the models presented in Table \ref{tab:comparison_studies} would have to be benchmarked on a single selected dataset.

\begin{table*}[]
\small
\begin{tabular}{lllll}
Model      & Study                           & Dataset                                                                & Nr. of classes & Accuracy                         \\ \hline
MARINE G14 & Present thesis & Coral reef: predator attacks within small fish on a coral reef  & Binary         & 81.53\%                          \\
MARINE G14 & Present thesis & AK fish: predation-related videos of fish from diverse settings & Binary         & 94.86\%                          \\ \hline
MAROON     & Schindler et al. \cite{schindler_action_2024}                & Deer videos in a forest environment from camera traps                  & 11             & 69.16\% (top-1) \\
DSRN       & Måløy et al. \cite{maloy_spatio-temporal_2019}               & Salmon in a cage at a farming site (either during feeding or not)      & Binary         & 80.0\%                           \\
Tiny VGG   & Feng et al. \cite{feng_action_2021}                          & Videos of tigers, lions and leopards in natural environments           & 3              & 92\% (average)                   \\ \hline
\end{tabular}
\caption{Comparison of MARINE AR outcomes on the coral reef and AK fish datasets to other animal AR studies conducted on video samples similarly specific to some other species and habitats.}
\label{tab:comparison_studies}
\end{table*}

Concerning the AD task, MARINE achieves significantly lower performance, with 0.00\% average precision in case of a 50\% t-IoU threshold. Comparing this with the outcome of the most similar study found during literature review, Nishioka et al.'s paper \cite{nishioka_detecting_2023} on action detection in recordings of zoo elephants, which reports 85.4\% average precision calculated on a frame-by-frame basis, it becomes clear that there is extreme room for improvement. Although the outcome with a threshold of 25\% t-IoU does come close to this (80.78\% average precision for MARINE G14), the lowered threshold questions the value of this metric. At the same time, it must be pointed out that VideoMAE achieves only 34.89\% average precision even with the lower threshold, which indicates that there could be comparative merit in the MARINE model for AD tasks as well. Furthermore, the difference between the dataset in \cite{nishioka_detecting_2023} and the coral reef dataset must be pointed out, as the former comprises recordings of elephants in zoo enclosures, and the latter fish in a coral reef, which could be a more challenging setting for action detection as the pose of the actors (small fish) is more difficult to discern. For a more conclusive comparison, both models would have to be tested on both of these datasets.

\subsection{Technical limitations}
There are several aspects in which MARINE could further be improved. For one, the processing of the training video samples could include data augmentation steps to mitigate the detrimental effects of small sample sizes and high class imbalances. This could ensure more robust generalization over the test sample, especially in case of highly specific video samples recorded in a single location, such as in the coral reef dataset.

Secondly, as the contribution of the motion-based frame selection method is inconsistent across the two datasets and different DINOv2 backbones (see Table \ref{tab:ablation_results}), a more sophisticated frame selection method could be applied to enhance results. At the same time, this must be done with caution in terms of computational efficiency: in case of more distinctive videos such as in the AK fish dataset, evenly spaced frames could provide an efficient alternative to ones selected through an informed method.

Furthermore, DINOv2 feature extraction could be utilized in ways other than using concatenated classifier tokens of entire frames to represent a video. For instance, identifying predator fish species as potential actors and extracting features on cutouts of the frames showing only the actor and its immediate environment (similarly to approaches in \cite{schindler_action_2024} and \cite{nishioka_detecting_2023}) may plausibly lead to enhanced performance.

Additionally, harnessing other modalities than visuals may prove highly beneficial in video understanding tasks, as demonstrated for audio in \cite{gao_listen_2020} and textual cues in MSQNet \cite{mondal_actor-agnostic_2023}, the latter especially relevant as it is currently state-of-the-art with 73.10\% mAP in multi-label AR on Animal Kingdom. If, instead or in addition to DINOv2, a pre-trained foundation model was utilized in MARINE which is capable of producing textual embeddings on visual input, then not only could performance improve on the datasets at hand, but the model might also obtain zero-shot capabilites for previously unseen videos.

\subsection{Applicability to different animal domains}
To understand the contribution of the study's results, the unique characteristics of the coral reef and AK fish datasets must be considered. Importantly, both of these datasets are small in computer vision terms and comprise videos of a selected animal group (fish) and a single target action class (predator attacks).

On the one hand, such specific samples are not uncommon in animal studies (see most animal AR papers discussed in Section \ref{sec:related_work}) and understandable considering that the goal of animal video understanding is often to study a certain behaviour in selected species. On the other hand, the small size and narrow scope of these samples makes it difficult to generalize study outcomes over a wider domain. In this regard, it is interesting to note that all models tested in this study tended to perform better on the AK fish dataset than on coral reef (see Table \ref{tab:AR_results}) even though the former comprised more species and environments. This highlights the fact that coral reef might be a generally more challenging dataset, especially because the shape and pose of actors (small fish in front of a dense coral background) is less discernible than those in Animal Kingdom.

Furthermore, the results of the multi-label setting experiment (Section \ref{sec:results_AK-sample}) show that there is room for improvement in MARINE's scalability to larger datasets with more diverse species and environments, especially if the action classes have a long-tailed distribution, as in Animal Kingdom \cite{ng_animal_2022}. In fact, the high imbalance between classes might be one of the reasons contributing to MARINE G14's comparatively low performance (23.79\% mAP against MSQNet's 73.10\%) on a representative sample of Animal Kingdom, which suggests that techniques specifically targeting this issue could help improve performance. For instance, in \cite{ng_animal_2022}, Ng et al. demonstrate that loss functions tailored to mitigate the effects of class imbalance can improve AR results on Animal Kingdom and might be a viable solution for MARINE as well.

In conclusion, while MARINE's performance, as reported in Section \ref{sec:results}, confirms that the model is a viable choice for AR tasks in the domain of fish predation videos, further work appears to be needed to adapt it to multi-label datasets which comprise more varied species, actions, and environments.
\section{Conclusion}
\label{sec:conclusion}

The study of encounters between predators and prey is of importance in ecosystem dynamics. This thesis investigates the extent to which computer vision techniques could be used to detect such rare ecological events, specifically predator attacks, in videos of fish species. The model MARINE is proposed, which comprises motion-based frame selection, feature extraction of selected frames using DINOv2 backbones, and a lightweight trainable classifier head to label video clips as containing a predator attack or not.

In the action recognition task on the coral reef and AK fish datasets, MARINE G14 is found to outperform state-of-the-art benchmark VideoMAE in terms of accuracy and F1-score on both the small and specific coral reef dataset (by close to 30\% points and 15\% points, respectively), and the larger and more diverse AK fish dataset (by close to 12\% points and 35\% points, respectively). Additionally, it is confirmed that larger LVM backbones (DINOv2 ViT-G/14 against ViT-S/14) increase performance in the AR task (by over 60\% points in accuracy for the coral reef, and 13\% points for the AK fish dataset), while register tokens do not prove to be of particular advantage. Similarly, the contribution of the motion-based frame selection method remains unclear, as it does not consistently improve MARINE's performance in an ablation study. Additionally, a multi-label AR experiment on a representative sample of Animal Kingdom reveals that MARINE cannot exceed the current state-of-the-art on a more diverse animal dataset, but it does perform comparably to other benchmarked methods (23.79\% mAP compared to existing benchmarks between 16.48\% and 73.10\% \cite{ng_animal_2022, mondal_actor-agnostic_2023}).  In terms of the AD task on the coral reef dataset, MARINE achieves rather low performance (0.00\% AP at 50\% t-IoU and 80.78\% AP at 25\% t-IoU), but still higher than benchmark model VideoMAE (34.89\% AP at 25\% t-IoU).

These results must be taken in light of the limitations introduced by the small size and narrow scope of the coral reef and AK fish datasets, as well as some technical aspects of MARINE with room for improvement. Consequently, future research could be directed towards two general areas. Firstly, MARINE's performance, especially in the AD task or in a multi-label setting, could be enhanced by applying more sophisticated frame selection methods, frame handling with focus on potential actors (e.g., feature extraction on cutouts of target animals), and foundation models utilizing non-visual modularities, such as textual cues. Secondly, MARINE could be tested on datasets specific to new animal domains, possibly a range of similarly small and narrow ones, to assess its suitability to be used in varied animal studies focusing on specific behaviours. These directions could each bring the state of research closer to a widely applicable model for identifying rare ecological events.

In conclusion, MARINE is a convincing starter model for modular computer vision frameworks applied to animal video understanding tasks. By expanding it, future research could bring much-needed further insight into the field of animal computer vision problems. This thesis hopes to offer a valuable contribution to this field and an expansion to the range of tools available for a better understanding of our natural ecosystems.
{
    \small
    \bibliographystyle{ieeenat_fullname}
    \bibliography{main}
}
\onecolumn

\appendix

\section{Description of the AK fish subset in Animal Kingdom}
\label{sec:apx:AK_fish_description}

\subsection{Predation-related actions in AK fish}
In the AK fish dataset, the following actions are considered predation-related, and videos in which at least one of these actions is performed by a fish species receive the positive ("attack") target label.

\begin{itemize}
    \item \textbf{Attacking:} Animal rushes and leaps on another another with bites, pecks or kicks.
    \item \textbf{Chasing:} Animal "runs" after the fleeing animal, however no biting occurs.
    \item \textbf{Fighting:} Include wrestling.
    \item \textbf{Fleeing:} Different from escaping, animal 'runs' away from its predator or danger, often with rapid change of direction but without having been caught first.
    \item \textbf{Retaliating:} Animal makes a defending attack (i.e., "attacks" its attacker back). Include ramming by goats.
    \item \textbf{Struggling:} Animal struggles from the clutch, grip, or bite by its predator.
    \item \textbf{Biting:} Animal sinks its teeth into the object or animal, but does not eat / feed / chew on the object or food.
    \item \textbf{Being eaten:} \textit{no description given in Animal Kingdom metadata.}
\end{itemize}

\begin{figure}
  \centering
  \includegraphics[width=0.4\textwidth]{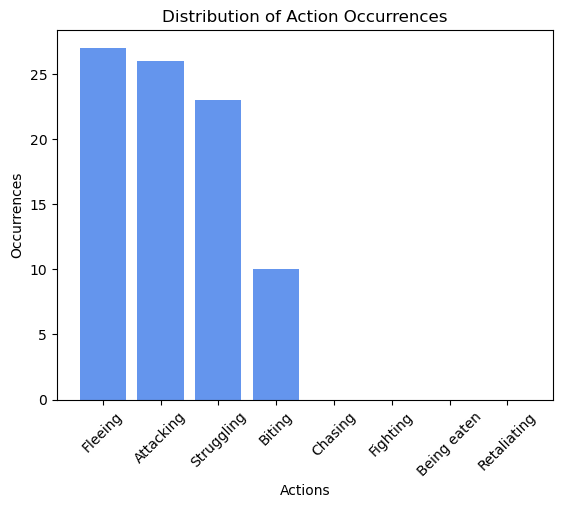}
  \caption{Frequency of predation-related actions labels in the AK fish dataset. One video may potentially include several of these actions.}
  \label{fig:apx:attack_action_dist}
\end{figure}

The frequency of these actions in the AK fish dataset (out of a total of 887 videos) is shown in Figure \ref{fig:apx:attack_action_dist}.

\section{Training MARINE}
\label{sec:apx:training}

This Appendix details the specifics of MARINE's training process. The complete implementation can be found on the thesis's GitHub page.

\subsection{Image transformations}
When handling video frames, the following transformations were applied during MARINE's training process.

\textbf{In the frame selection module:}
\begin{enumerate}
    \item During motion-based frame selection, the frames are converted to greyscale images to calculate the dissimilarity scores between consecutive frames.

\end{enumerate}

\textbf{In the feature extraction module:}
\begin{enumerate}
    \item The shorter side of the video frames is resized to 448 pixels.
    \item The longer side of the video frames is resized to preserve the aspect ratio. If the new length is not a multiple of the DINOv2 backbone's patch size (14 pixels), then it is re-set using the following formula: $new\_height = \lfloor new\_height / patch\_size \rfloor * patch\_size$, resulting in the (very slight) distortion of the image.
    \item The video frames are represented as RGB images (i.e., three-dimensional tensors).
\end{enumerate}

\subsection{Class weights}
As high imbalance is observed both in the coral reef and AK fish datasets, class weights are used to calculate the loss during training. Class weights are calculated using the formula:

$$w_k = \frac{n}{K * f_k}$$,

where $w_k$ is the weight of class $k$, $n$ is the total number of samples, $K$ is the total number of classes, and $f_k$ is the number of samples in class $k$.

\subsection{Cross-validation specifics}

Table \ref{tab:apx:cross_val} summarizes the hyperparameters tuned during 3-fold cross-validation of MARINE's classification head, as well as the best parameters found for MARINE G14 in case of the coral reef and AK fish datasets. As the rarity of predator attacks (i.e., high class imbalance) necessitates good ability to identify these events without compromising on specificity (or, conversely, the false positive rate), the ROC AUC value of the classifiers was used to select the best model during cross validation. Additional settings for the training process are shown in Table \ref{tab:apx:training}.

Notably, the MARINE models were trained for 10 epochs during cross-validation, but only for a single epoch on the whole training set. The reason for this choice is that increasing the number of epochs (to 10) on the whole training set results in decreased performance on the test set for most MARINE configurations, and only a slight increase for some of them. The experiments may be reproduced with any selected number of epochs using the implementation scripts provided on the project's GitHub page.

\begin{table}[]
\centering
\begin{tabular}{llll}
\multirow{2}{*}{Hyperparameter}                & \multirow{2}{*}{Values tested during cross-validation} & \multicolumn{2}{l}{Best value for MARINE G14} \\
                                               &                                                        & Coral reef              & AK fish             \\ \hline
Hidden dense layers with ReLu (128 nodes each) & 0 to 3                                                 & 2                       & 1                   \\
Dropout rate                                   & 0.0, 0.25, 0.5                                         & 0.25                    & 0.0                 \\
Learning rate                                  & 0.01, 0.001, 0.0001                                    & 0.001                   & 0.0001              \\ \hline
\end{tabular}
\caption{Hyperparameters selected through cross-validation for MARINE's classification head. The best parameters found for MARINE G14 (without registers) are reported for the coral reef and AK fish datasets.}
\label{tab:apx:cross_val}
\end{table}

\begin{table}[]
\centering
\begin{tabular}{ll}
Parameter     & Value                \\ \hline
Batch size    & 32                   \\
Epochs        & 10 (cross-validation), 1 (final training)                   \\
Optimizer     & Adam                 \\
Loss function & Binary cross-entropy \\ \hline
\end{tabular}
\caption{Training parameters used during MARINE's cross-validation process and for training the classification head with the selected hyperparameters on the complete train set for final testing.}
\label{tab:apx:training}
\end{table}

\subsection{Selecting the best threshold}
In the binary classification setting of the AR task, the best threshold to label a sample as predicted positive based on the sigmoid output of the classifier head is determined by optimizing for F1 score on the training set. This means that the F1 score is calculated using each of the unique sigmoid outputs as the threshold for a positive prediction, and the threshold which maximizes the score is selected. This is the threshold which is then used to classify the samples in the test set.

\subsection{Computational resources}

Table \ref{tab:apx:resources} shows the computational resources needed to train the largest model proposed in this thesis, MARINE G14 (without registers), on the largest dataset employed in this study, the representative sample of Animal Kingdom, comprising 1000 videos. Resources are shown for each of MARINE's modules separately. The training process used to obtain the resource estimates was executed using L4 GPUs on Google Colab and is described in detail in Section \ref{sec:method_experiments_AK-sample} and Appendix \ref{sec:apx:AK-sample}.

\begin{table}[]
\centering
\begin{tabular}{lll}
MARINE G14 module            & Colab computing units & Time (hours) \\ \hline
Motion-based frame selection &                       1.5&              0.12\\
DINOv2 feature extraction    & 20                    & 9.5          \\
Training classifier head     & 2.7& 0.3\\ \hline
\end{tabular}
\caption{Computing units and time needed to execute the modules of MARINE G14 (without registers) on Google Colab L4 GPUs on the representative sample of Animal Kingdom, comprising 1000 videos.}
\label{tab:apx:resources}
\end{table}

\section{Fine-tuning VideoMAE}
\label{sec:apx:VideoMAE}

The classification head of the VideoMAE model with a ViT-B backbone and pre-trained on Kinetics-400 was fine-tuned on the coral reef and AK fish datasets. For each video, 10 frames are selected using the motion-based method (as in MARINE), resized to 448x448 pixels, and pixel values are normalized. Besides the specifics mentioned in Section \ref{sec:method_experiments_AR}, Table \ref{tab:apx:VideoMAE} summarizes the settings of the training process. VideoMAE is fine-tuned on the AK fish dataset with a batch size of 8 for 4 epochs, as experiments with higher values (batch size of 32 for 10 epochs) resulted in decreased performance on this dataset. The complete implementation can be found on the project's GitHub page.

\begin{table}[]
\centering
\begin{tabular}{ll}
Parameter     & Value                              \\ \hline
Batch size    & 32 (coral reef), 8 (AK fish)\\
Learning rate & 0.001                              \\
Epochs        & 10 (coral reef), 4 (AK fish)\\
Optimizer     & Adam                               \\
Loss function & Binary cross entropy (with logits) \\ \hline
\end{tabular}
\caption{Training parameters used during VideoMAE's fine-tuning process on the coral reef and AK fish datasets.}
\label{tab:apx:VideoMAE}
\end{table}

\section{Registers in the MARINE models' DINOv2 backbones}
\label{sec:apx:registers}

\begin{figure}[ht]
    \centering
    \begin{subfigure}[b]{0.45\textwidth}
        \centering
        \includegraphics[width=\textwidth]{figures/ROC_MARINEG14_woRegisters.png}
    \end{subfigure}
    \begin{subfigure}[b]{0.45\textwidth}
        \centering
        \includegraphics[width=\textwidth]{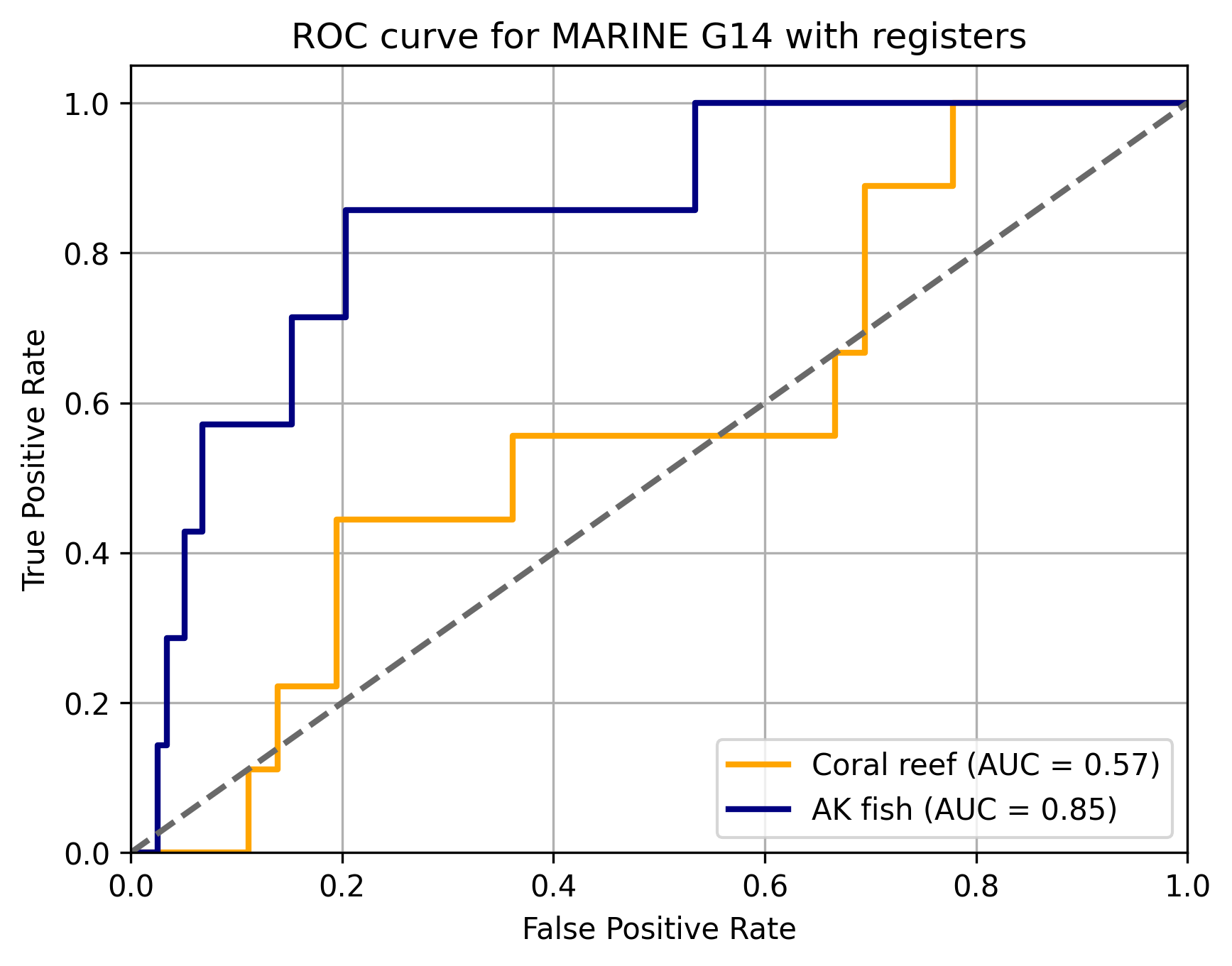}
    \end{subfigure}
    \caption{ROC curves of MARINE G14 with and without the use of registers, on both the coral reef (orange curve) and AK fish (blue curve) datasets.}
    \label{fig:apx:AR_registers}
\end{figure}

Table \ref{tab:apx:AR_registers} shows the results of the AR task on the coral reef and AK fish datasets for all configurations of the MARINE model: with ViT-S/14 or ViT-G/14 DINOv2 backbones, and with or without registers. Additionally, Figure \ref{fig:apx:AR_registers} shows the ROC curves of MARINE G14 obtained with and without the use of registers. As can be seen, the use of register tokens does not consistently improve the performance of MARINE although it does appear to carry some advantage on the AK fish dataset (in terms of the performance metrics shown in Table \ref{tab:apx:AR_registers} for MARINE S14, and in terms of the ROC AUC for MARINE G14).

\begin{table}[]
\begin{tabular}{llcllll}
Dataset    & Model      & Registers & Accuracy (\%)  & Recall (\%)    & Precision (\%) & F1-score(\%)   \\ \hline
Coral reef & MARINE S14 & \xmark & 19.51 +/ 12.25 & 100.00 +/ 0.00 & 19.51 +/ 12.25 & 32.19 +/ 17.30 \\
Coral reef & MARINE S14 & \cmark      & 19.51 +/ 12.25 & 100.00 +/ 0.00 & 19.51 +/ 12.25 & 32.19 +/ 17.30 \\ \hline
Coral reef & MARINE G14 & \xmark     & \textbf{81.53 +/ 10.36} & \textbf{52.52 +/ 33.77} & \textbf{53.01 +/ 37.24} & \textbf{51.06 +/ 29.60} \\
Coral reef & MARINE G14 & \cmark      & 73.71 +/ 14.31 & 44.62 +/ 37.35 & 36.06 +/ 33.34 & 38.71 +/ 32.43 \\ \hline
AK fish    & MARINE S14 & \xmark     & 81.74 +/ 7.05  & 57.40 +/ 41.50 & 16.07 +/ 14.68 & 24.62 +/ 20.54 \\
AK fish    & MARINE S14 & \cmark      & \textbf{88.38 +/ 5.35}  & \textbf{59.39 +/ 40.18} & \textbf{25.24 +/ 22.92} & \textbf{34.41 +/ 25.97} \\ \hline
AK fish    & MARINE G14 & \xmark     & \textbf{94.86 +/ 3.97}  & \textbf{63.31 +/ 41.60} & \textbf{51.64 +/ 36.86} & \textbf{55.14 +/ 33.18} \\
AK fish    & MARINE G14 & \cmark      & 91.31 +/ 5.27  & 56.41 +/ 40.99 & 32.32 +/ 25.36 & 39.98 +/ 27.93 \\ \hline
\end{tabular}
\caption{Performance of the MARINE models with and without registers. The best results in terms of each metric are highlighted in bold for each pair of MARINE configurations with and without registers on the same dataset.}
\label{tab:apx:AR_registers}
\end{table}

\section{Multi-label action recognition on Animal Kingdom sample}
\label{sec:apx:AK-sample}

\subsection{Representative sampling}
A representative sample is taken from Animal Kingdom by randomly selecting 1000 videos while ensuring that the proportion of test instances (as defined in the Animal Kingdom dataset) remains the same as in the original dataset, resulting in 797 train and 203 test instances being sampled.

The sample dataset contains 93 classes of the original 140, with some low-frequency classes missing (due to the long-tailed distribution of Animal Kingdom emphasized by \cite{ng_animal_2022}). A chi-square test of homogeneity with p-value = 0.68 confirms that the classes in the sample follow the same distribution as in the original Animal Kingdom dataset.

\subsection{Adapting MARINE for multi-label classification}
The videos in Animal Kingdom may simultaneously possess several action labels. To adapt MARINE for this multi-label task, the following changes are implemented in the training process (all other settings remain identical to the binary classification setup).

\begin{itemize}
    \item Instead of class weights, sample weights are used when training the classifier head on the whole training set (after cross-validation). Sample weights are calculated as

$$s_i = max_{k \in L_i}w_k$$
where $s_i$ is the weight of a sample $s$, $L_i$ is the set of classes which $s$ belongs to, and $w_k$ is the weight of class $k$ calculated as 
    $$w_k = \frac{n}{K * f_k + \epsilon}$$
where $n$ is the total number of samples, $K$ is the number of classes, $f_k$ is the number of samples in class $k$, and $\epsilon$ is a smoothing parameter set to $10^{-6}$.

    \item Instead of a single sigmoid output, the last layer of the classifier head now includes one node with sigmoid activation for each class in the representative sample (93 overall).

    \item Instead of the ROC AUC value, the best model hyperparameters are now selected to optimize for mAP during cross-validation. The same hyperparameter settings are tested as in the binary AR setting, and the same training settings are used (as shown in Tables \ref{tab:apx:cross_val} and \ref{tab:apx:training}, except for the number of epochs, which is described below).

    \item Instead of a single epoch for training on the complete training set (as for the binary AR training process described in Appendix \ref{sec:apx:training}), MARINE G14 is trained for 10 epochs on the Animal Kingdom sample, as, contrary to the binary AR task, this results in considerably increased performance in this case (from 5.09\% mAP to 23.79\% mAP).

    \item The best threshold for the sigmoid output to consider an instance as predicted positive for some class is determined by testing thresholds between 0.1 and 0.9 (with a 0.1 step), and selecting the one which produces the highest micro-averaged F1-score. The selected threshold is applied to every class in the dataset.
    
\end{itemize}

The best hyperparameter settings found for MARINE G14 (without registers) in a multi-label setting are presented in Table \ref{tab:apx:multilabel_parameters}.

\begin{table}[]
\centering
\begin{tabular}{ll}
Hyperparameter                                 & Value \\ \hline
Number of dense hidden layers (128 nodes each) & 2\\
Drop-out rate                                  & 0.0\\
Learning rate                                  & 0.01\\
Threshold for positive predictions             & 0.4\\  \hline
\end{tabular}
\caption{Best hyperparameter settings found through cross-validation for MARINE G14 (without registers) in the multi-label setting on the representative sample of Animal Kingdom.}
\label{tab:apx:multilabel_parameters}
\end{table}

\end{document}